\documentclass[final]{cvpr}

\usepackage{times}
\usepackage{epsfig}
\usepackage{graphicx}
\usepackage{amsmath}
\usepackage{amssymb}

\usepackage{bm}

\usepackage{amsthm}
\usepackage{amsbsy}
\usepackage{booktabs}
\usepackage{enumitem}
\usepackage{mathtools}
\usepackage{subfigure}
\usepackage{tabularx}
\usepackage{multirow}
\usepackage{caption}
\usepackage{overpic}
\usepackage{xcolor}
\usepackage{comment}
\usepackage[title]{appendix}
\usepackage{chngcntr}
\usepackage{flushend}
\usepackage{balance}

\DeclarePairedDelimiterX{\infdivx}[2]{(}{)}{%
  #1\;\delimsize\|\;#2%
}
\DeclarePairedDelimiterX{\inp}[2]{\langle}{\rangle}{#1, #2}

\DeclarePairedDelimiter{\norm}{\big\lVert}{\big\rVert}
\DeclarePairedDelimiter{\bignorm}{\Big\lVert}{\Big\rVert}
\DeclareMathOperator*{\argmin}{\arg\!\min}

\newlength{\oldparindent}
\newcommand{\app}{SCALE\xspace}

\renewcommand{\eg}{e.g.~}

\renewcommand{\paragraph}[1]{\vspace{5pt}{\noindent\textbf{#1}}}

\newcommand\blfootnote[1]{%
  \begingroup
  \renewcommand\thefootnote{}\footnote{#1}%
  \addtocounter{footnote}{-1}%
  \endgroup
}

\newcommand{\descriptorMath}{$\mathbf{u}_k$\xspace}

\newcommand{\localfeatMath}{$\bm{z}_k$\xspace}

\newcommand*{\affaddr}[1]{#1} 
\newcommand*{\affmark}[1][*]{\textsuperscript{#1}}
\newcommand*{\email}[1]{\small{\texttt{#1}}}

\usepackage[pagebackref=true,breaklinks=true,colorlinks,bookmarks=false]{hyperref}

\def\cvprPaperID{428} 
\def\confYear{CVPR 2021}

\begin{document}

\title{SCALE: Modeling Clothed Humans with a \\Surface Codec of Articulated Local Elements}
\author{
Qianli Ma\affmark[1,2]\quad
Shunsuke Saito\affmark[1]\textsuperscript{$\dagger$} \quad
Jinlong Yang\affmark[1] \quad
Siyu Tang\affmark[2]\quad
Michael J. Black\affmark[1]
\\
\affaddr{\affmark[1]Max Planck Institute for Intelligent Systems, T\"ubingen, Germany} \quad \affaddr{\affmark[2]ETH Z\"urich}\\
\email{\{qma,ssaito,jyang,black\}@tuebingen.mpg.de},\quad \email{\{siyu.tang\}@inf.ethz.ch}
}
\twocolumn[{%
\renewcommand\twocolumn[1][]{#1}%
\maketitle
\begin{center}
    \newcommand{\teaserwidth}{\textwidth}
\vspace{-0.2in}
    \centerline{
    \includegraphics[width=1\linewidth]{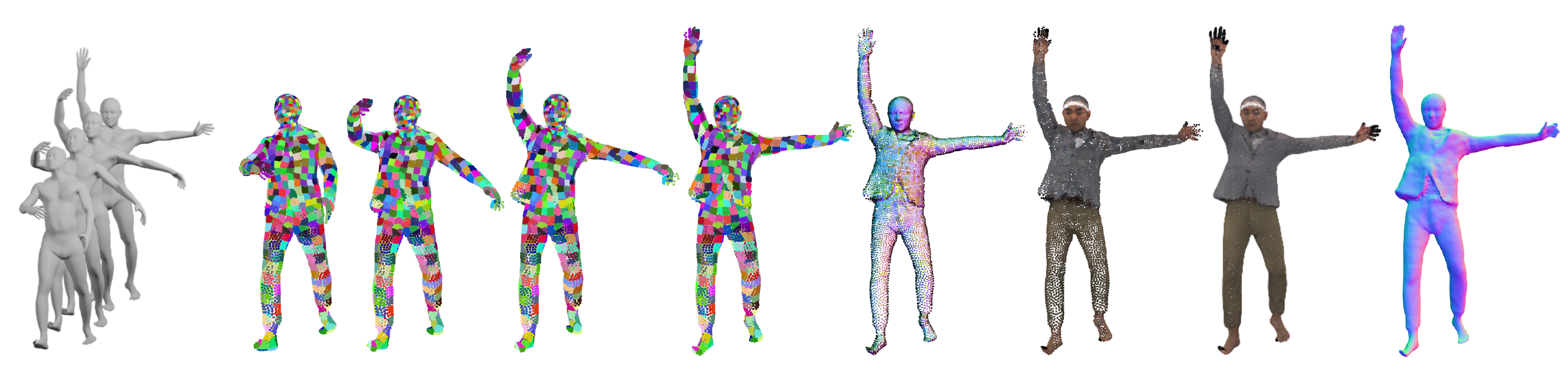}
    \put(-488,-8){\small{Query poses}}
    \put(-400,-8){\small{Predicted Articulated Local Elements}}
    \put(-233,-8){\small{with normals}}
    \put(-175,-8){\small{with texture}}
    \put(-125,-8){\small{neural rendered}}
    \put(-56,-8){\small{meshed}}
     }
  \vspace{3pt}
  \captionof{figure}{\textbf{SCALE.} Given a sequence of posed but minimally-clothed 3D bodies, \app predicts 798 articulated surface elements (patches, each visualized with a unique color) to dress the bodies with realistic clothing that moves and deforms naturally even in the presence of topological change. The resulting dense point sets have correspondence across different poses, as shown by the consistent patch colors. The result also includes predicted surface normals and texture, with which the point cloud can either be turned into a 3D mesh or directly rendered as realistic images using neural rendering techniques.}
\label{fig:teaser}
\end{center}%
}]

\maketitle
\thispagestyle{empty} 
\begin{abstract}

Learning to model and reconstruct humans in clothing is challenging due to articulation, non-rigid deformation, and varying clothing types and topologies. 
To enable learning, the choice of representation is the key.
Recent work uses neural networks to parameterize local surface elements.
This approach captures locally coherent geometry and non-planar details, can deal with varying topology, and does not require registered training data.
However, naively using such methods to model 3D clothed humans fails to capture fine-grained local deformations and generalizes poorly. 
To address this, we present three key innovations: 
First, we deform surface elements based on a human body model such that large-scale deformations caused by articulation are explicitly separated from topological changes and local clothing deformations. 
Second, we address the limitations of existing neural surface elements by regressing local geometry from local features, significantly improving the expressiveness.
Third, we learn a pose embedding on a 2D parameterization space that encodes posed body geometry, improving generalization to unseen poses by reducing non-local spurious correlations.
We demonstrate the efficacy of our surface representation by learning models of complex clothing from point clouds. The clothing can change topology and deviate from the topology of the body. Once learned, we can animate previously unseen motions, producing high-quality point clouds, from which we generate realistic images with neural rendering.
We assess the importance of each technical contribution and show that our approach outperforms the state-of-the-art methods in terms of reconstruction accuracy and inference time. The code is available for research purposes at {\small\url{https://qianlim.github.io/SCALE}}.
\blfootnote{$\dagger$ Now at Facebook Reality Labs.}
\end{abstract}

\vspace{-10pt}

\section{Introduction}
\label{sec:intro}
While models of humans in clothing would be valuable for many tasks in computer vision such as body pose and shape estimation from images and videos~\cite{bogo2016smplify,ExPose:2020,kanazawa2018end,kanazawa2019learning,VIBE:CVPR:2020,SPIN:ICCV:2019} and synthetic data generation~\cite{ranjan2019multi,ranjan2018learning,varol2017learning,PLACE:3DV:2020}, most existing approaches are based on ``minimally-clothed'' human body models~\cite{anguelov2005scape,joo2018total,loper2015smpl, STAR:ECCV:2020, pavlakos2019expressive,xu2020ghum}, which do not represent clothing. 
To date, statistical models for clothed humans remain lacking despite the broad range of potential applications. 
This is likely due to the fact that modeling 3D clothing shapes is much more difficult than modeling body shapes.
Fundamentally, several characteristics of clothed bodies present technical challenges for representing clothing shapes.

The first challenge is that clothing shape varies at different spatial scales driven by global body articulation and local clothing geometry. The former requires the representation to properly handle human pose variation, while the latter requires local expressiveness to model folds and wrinkles.
Second, a representation must be able to model smooth cloth surfaces and also sharp discontinuities and thin structures. 
Third, clothing is diverse and varies in terms of its topology. 
The topology can even change with the motion of the body.
Fourth, the relationship between the clothing and the body changes as the clothing moves relative to the body surface.
Finally, the representation should be compatible with existing body models and should support fast inference and rendering, enabling real-world applications.

Unfortunately, none of the existing 3D shape representations satisfy all these requirements. 
The standard approach uses 3D meshes that are draped with clothing using physics simulation \cite{baraff1998large,liang2019differentiable,liu2013fast}. 
These require manual clothing design and the physics simulation makes them inappropriate for inference.
Recent work starts with classical rigged 3D meshes and blend skinning but uses machine learning to model clothing shape and local non-rigid shape deformation. 
However, these methods often rely on pre-defined garment templates~\cite{bhatnagar2019mgn,lahner2018deepwrinkles,CAPE:CVPR:20, patel20tailornet}, and the fixed correspondence between the body and garment template restricts them from generalizing to arbitrary clothing topology. 
Additionally, learning a mesh-based model requires registering a common 3D mesh template to scan data. This is time consuming, error prone, and limits topology change~\cite{pons2017clothcap}.
New neural implicit representations \cite{chen2018implicit_decoder,mescheder2019occupancy,park2019deepsdf}, on the other hand, are able to reconstruct topologically varying clothing types \cite{chibane2020implicit,corona2021smplicit,saito2020pifuhd}, but are not consistent with existing graphics tools, are expensive to render,
and are not yet suitable for fast inference. 
Point clouds are a simple representation that also supports arbitrary topology \cite{fan2017point, lin2018learning, yang2019pointflow} and does not require data registration, but highly detailed geometry requires many points.

A middle ground solution is to utilize a collection of parametric surface elements that smoothly conform to the global shape of the target geometry~\cite{deprelle2019learning,groueix2018atlasnet,yang2018foldingnet,yuan2018pcn,zhao20193d}. As each element can be freely connected or disconnected, topologically varying surfaces can be effectively modeled while retaining the efficiency of explicit shape inference. 
Like point clouds, these methods can be learned without data registration.

However, despite modeling coherent global shape, existing surface-element-based representations often fail to generate local structures with high-fidelity. The key limiting factor is that shapes are typically decoded from \textit{global} latent codes~\cite{groueix2018atlasnet,yang2018foldingnet,yuan2018pcn}, i.e.~the network needs to learn both the global shape statistics (caused by articulation) and a prior for local geometry (caused by clothing deformation) at once. 
While the recent work of~\cite{groueix20183d} shows the ability to handle articulated objects, these methods often fail to capture local structures such as sharp edges and wrinkles, hence the ability to model \textit{clothed} human bodies has not been demonstrated. 

In this work, we extend the surface element representation to create a clothed human model that meets \textit{all} the aforementioned desired properties. 
We support articulation by defining the surface elements on top of a minimal clothed body model. 
To densely cover the surface, and effectively model local geometric details, we first introduce a global patch descriptor that differentiates surface elements at different locations, enabling the modeling of hundreds of local surface elements with a single network, and then regress local non-rigid shapes from local pose information, producing folding and wrinkles. Our new shape representation, \textit{Surface Codec of Articulated Local Elements}, or SCALE, demonstrates state-of-the-art performance on the challenging task of modeling the per-subject pose-dependent shape of clothed humans, setting a new baseline for modeling topologically varying high-fidelity surface geometry with explicit shape inference. See Fig.~\ref{fig:teaser}.

In summary, our contributions are:
(1) an extension of surface element representations to non-rigid articulated object modeling;
(2) a revised local elements model that generates local geometry from local shape signals instead of a global shape vector;
(3) an explicit shape representation for clothed human shape modeling that is robust to varying topology, produces high-visual-fidelity shapes, is easily controllable by pose parameters, and achieves fast inference;
and (4) a novel approach for modeling humans in clothing that does not require registered training data and generalizes to various garment types of different topology, addressing the missing pieces from existing clothed human models.
We also show how neural rendering is used together with our point-based representation to produce high-quality rendered results.
The code is available for research purposes at {\small\url{https://qianlim.github.io/SCALE}}.

\section{Related Work}
\paragraph{Shape Representations for Modeling Humans.}
Surface meshes are the most commonly used representation for human shape due to their efficiency and compatibility with graphics engines. Not only human body models~\cite{anguelov2005scape, loper2015smpl, STAR:ECCV:2020, xu2020ghum} but also various clothing models leverage 3D mesh representations as separate mesh layers~\cite{danvevrek2017deepgarment,guan2012drape, gundogdu2019garnet, lahner2018deepwrinkles,patel20tailornet,santesteban2019} or displacements from a minimally clothed body~\cite{bhatnagar2019mgn,CAPE:CVPR:20, Neophytou2014layered,tiwari20sizer,xiang2020monoclothcap,yang2018physics}. 
Recent advances in deep learning have improved the fidelity and expressiveness of mesh-based approaches using graph convolutions~\cite{CAPE:CVPR:20}, multilayer perceptrons (MLP)~\cite{patel20tailornet}, and 2D convolutions~\cite{jin2018pixel, lahner2018deepwrinkles}. 
The drawback of mesh-based representations is that topological changes are difficult to model. 
Pan et al.~\cite{pan2019deep} propose a topology modification network (TMN) to support topological change, however it has difficulty
learning large topological changes from a single template mesh \cite{zhu2020deep}. 

To support various topologies, neural implicit surfaces~\cite{mescheder2019occupancy, park2019deepsdf} have recently been applied to clothed human reconstruction and registration~\cite{bhatnagar2020ipnet, saito2020pifuhd}. The extension of implicit surface models to unsigned distances~\cite{chibane2020ndf} or probability densities~\cite{ShapeGF} even allows thin structures to be represented with high resolution. 
Recent work~\cite{amoss2020igr} also shows the ability to learn a signed-distance function (SDF) directly from incomplete scan data. 
Contemporaneous with our work, SCANimate~\cite{SCANimate:2021}, learns an implicit shape model of clothed people from raw scans. 
Also contemporaneous is SMPLicit~\cite{corona2021smplicit}, which learns an implicit clothing representation
that can be fit to scans or 2D clothing segmentations.
Despite the impressive reconstruction quality of implicit methods, extracting an explicit surface is time consuming but necessary for many applications.

Surface element representations~\cite{deprelle2019learning, groueix2018atlasnet, williams2019deep, yang2018foldingnet} are promising for modeling clothing more explicitly. These methods approximate various topologies with locally coherent geometry by learning to deform single or multiple surface elements.
Recent work improves these patch-based approaches by incorporating differential geometric regularization \cite{bednarik2020, deng2020better}, demonstrating simple clothing shape reconstruction. Although patch-based representations relax the topology constraint of a single template by representing the 3D surface as the combination of multiple surface elements, reconstruction typically lacks details as the number of patches is limited due to the memory requirements, which scale linearly with the number of patches. 
While SCALE is based on these neural surface representations, we address the limitations of existing methods in Sec.~\ref{sec:ale}, enabling complex clothed human modeling.

\paragraph{Articulated Shape Modeling.}
Articulation often dominates large-scale shape variations for articulated objects, such as hands and human bodies. 
To efficiently represent shape variations, articulated objects are usually modeled with meshes driven by an embedded skeleton~\cite{anguelov2005scape, loper2015smpl, romero17mano}. 
Mesh-based clothing models also follow the same principle~\cite{guan2012drape, lahner2018deepwrinkles, CAPE:CVPR:20, Neophytou2014layered, patel20tailornet}, where shapes are decomposed into articulated deformations and non-rigid local shape deformations. 
While the former are explained by body joint transformations, the latter can be efficiently modeled in a canonical space. One limitation, however, is that registered data or physics-based simulation is required to learn these deformations on a template mesh with a fixed topology. 
In contrast, recent work on articulated implicit shape modeling~\cite{deng2019neural,LEAP:CVPR:21,SCANimate:2021} does  not require surface registration. In this work we compare with Deng et al.~\cite{deng2019neural} on a clothed human modeling task from point clouds and show the superiority of our approach in terms of generalization to unseen poses, fidelity, and inference speed. 

\paragraph{Local Shape Modeling.}
Instead of learning 3D shapes solely with a global feature vector, recent work shows that learning from local shape variations leads to detailed and highly generalizable 3D reconstruction~\cite{chabra2020deep, chibane2020implicit,genova2020local,Paschalidou2021CVPR,Peng2020ECCV,saito2019pifu,tretschk2020patchnets}. Leveraging local shape priors is effective for 3D reconstruction tasks from 3D point clouds~\cite{chabra2020deep, chibane2020implicit,genova2020local,jiang2020local,Peng2020ECCV, tretschk2020patchnets} and images~\cite{Paschalidou2021CVPR,saito2019pifu,xu2019disn}. Inspired by this prior work, SCALE leverages both local and global feature representations, which leads to high-fidelity reconstruction as well as robust generalization to unseen poses.

\section{SCALE}
\begin{figure*}[tb]
    \centering
    \includegraphics[width=0.95\linewidth]{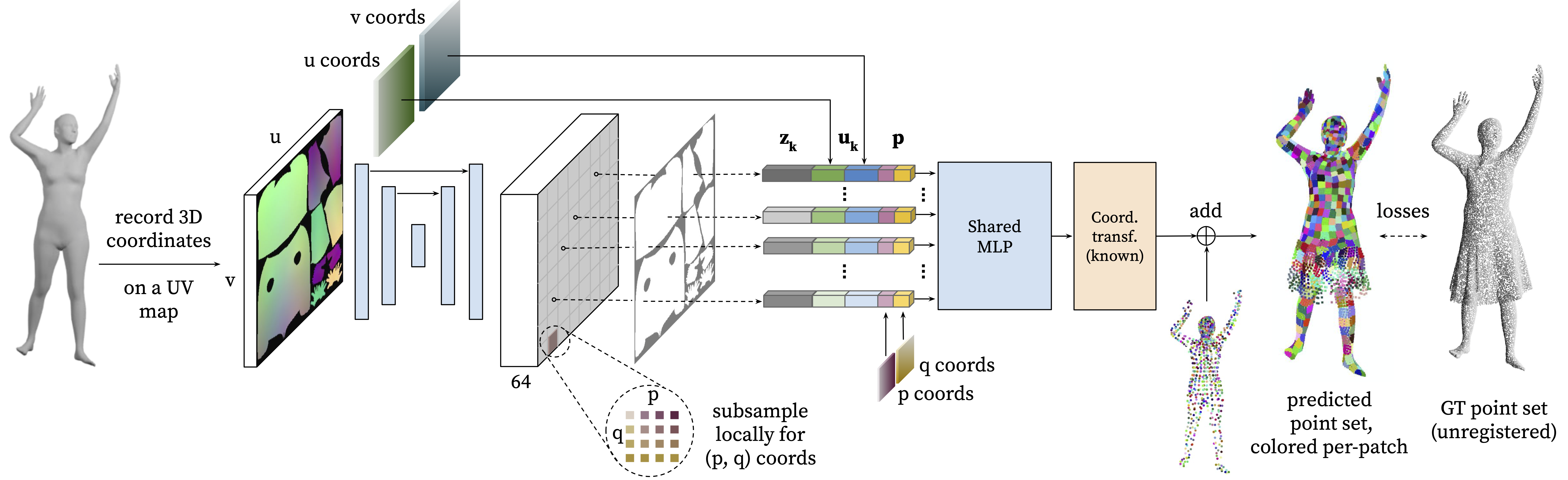}
    \vspace{-3pt}
    \caption{\small \textbf{Method overview.} 
    Given a posed, minimally-clothed body, we define a set of points on its surface and associate a local element (a square patch) with each of them.
    The body points' positions in $\mathbb{R}^3$ are recorded on a 2D UV positional map, which is convolved by a UNet to obtain pixel-aligned local pose features $\bm{z}_k$. 
    The 2D coordinates $\mathbf{u}_k = (u_k, v_k)$ on the UV map, the pose features $\bm{z}_k$, and the 2D coordinates within the local elements $\mathbf{p}=(p, q)$ are fed into a shared MLP to predict the deformation of the local elements in the form of residuals from the body. 
    The inferred local elements are finally articulated by the known transformations of corresponding body points to generate a posed clothed human. Each patch and its corresponding body point is visualized with the same color.}
    \label{fig:method_overview}
\end{figure*}

Figure~\ref{fig:method_overview} shows an overview of SCALE. Our goal is to model clothed humans with a topologically flexible point-based shape representation that supports fast inference and animation with SMPL pose parameters~\cite{loper2015smpl}.
To this end, we model pose-dependent shape variations of clothing using a collection of local surface elements (patches) that are associated with a set of pre-defined locations on the body.
Our learning-based local pose embedding further improves the generalization of pose-aware clothing deformations (Sec.~\ref{sec:ale}).
Using this local surface element representation, we train a model for each clothing type to predict a set of 3D points representing the clothed body shape given an unclothed input body.
Together with the predicted point normals and colors, the dense point set can be meshed or realistically rendered with neural rendering (Sec.~\ref{sec:losses}).

\subsection{Articulated Local Elements}\label{sec:ale}
\setlength{\abovedisplayskip}{5pt}
\setlength{\belowdisplayskip}{5pt}
While neural surface elements~\cite{groueix2018atlasnet, yang2018foldingnet, yuan2018pcn, zhao20193d} offer locally coherent geometry with fast inference, the existing formulations have limitations that prevent us from applying them to  clothed-human modeling. We first review the existing neural surface elements and introduce our formulation that addresses the drawbacks of the prior work.

\paragraph{Review: Neural Surface Elements.}
The original methods that model neural surface elements~\cite{groueix2018atlasnet, yang2018foldingnet} learn a function to generate 3D point clouds as follows:
\begin{equation}
\label{eq:atlas}
f_w(\mathbf{p}; \bm{z}): \mathbb{R}^D\times \mathbb{R}^\mathcal{Z} \to \mathbb{R}^3,
\end{equation}
where $f_w$ is a multilayer perceptron (MLP) parameterized by weights $w$, $\mathbf{p} \in \mathbb{R}^D$ is a point on the surface element, and $\bm{z} \in \mathbb{R}^\mathcal{Z}$ is a global feature representing object shape. 
Specifically, $f_w$ maps $\mathbf{p}$ on the surface element to a point on the surface of a target 3D object conditioned by a shape code $\bm{z}$. Due to the inductive bias of MLPs, the resulting 3D point clouds are geometrically smooth within the element \cite{williams2019deep}. While this smoothness is desirable for surface modeling, to support different topologies, AtlasNet~\cite{groueix2018atlasnet} requires multiple surface elements represented by individual networks, which increases network parameters and memory cost. The cost is linearly proportional to the number of patches. As a result, these approaches limit expressiveness for topologically complex objects such as clothed humans. 

Another line of work represents 3D shapes using a collection of local elements. 
PointCapsNet~\cite{zhao20193d} decodes a \textit{local} shape code $\{\bm{z}_k\}_{k=1}^K$, where $K$ is the number of local elements, into \textit{local} patches with separate networks:
\begin{equation}
\label{eq:3dcapsule}
f_{w_k}(\mathbf{p}; \bm{z}_k): \mathbb{R}^D\times \mathbb{R}^\mathcal{Z} \to \mathbb{R}^3.
\end{equation}
While modeling local shape statistics instead of global shape variations improves the generalization and training efficiency for diverse shapes, the number of patches is still difficult to scale up as in AtlasNet for the same reason. 

Point Completion Network (PCN)~\cite{yuan2018pcn} uses two-stage decoding: the first stage predicts a coarse point set of the target shape, then these points are used as basis points for the second stage. At each basis location $\bm{b}_k\in\mathbb{R}^3$, points $\mathbf{p}$ from a local surface element (a regular grid) are sampled and fed into the second decoder as follows:
\begin{equation}
\label{eq:pcn}
f_w(\bm{b}_k, \mathbf{p}; \bm{z}): \mathbb{R}^{3} \times \mathbb{R}^{D} \times \mathbb{R}^\mathcal{Z} \to \mathbb{R}^3.
\end{equation}
Notably, PCN utilizes a single network to model a large number of local elements, improving the expressiveness with an arbitrary shape topology. However, PCN relies on a global shape code $\bm{z}$ that requires learning global shape statistics, resulting in poor generalization to unseen data samples as demonstrated in Sec.~\ref{exp:clo_deform_local_global}.

\paragraph{Articulated Local Elements.}
For clothed human modeling, the shape representation needs to be not only expressive but also highly generalizable to unseen poses. These requirements and the advantages of the prior methods lead to our formulation:
\begin{equation}
\label{eq:scale}
g_w(\mathbf{u}_k, \mathbf{p} ; \bm{z}_k): \mathbb{R}^{D_1} \times \mathbb{R}^{D_2} \times \mathbb{R}^\mathcal{Z} \to \mathbb{R}^3,
\end{equation}
where $\mathbf{u}_k \in \mathbb{R}^{D_1}$ is a global patch descriptor that provides inter-patch relations and helps the network $g_w$ distinguish different surface elements, 
and $\mathbf{p} \in \mathbb{R}^{D_2}$ are the local (intra-patch) coordinates within each surface element. Importantly, our formulation achieves higher expressiveness by efficiently modeling a large number of local elements using a single network as in~\cite{yuan2018pcn} while improving generality by learning local shape variations with \localfeatMath. 
Moreover, unlike the existing methods \cite{groueix20183d,groueix2018atlasnet,yang2018foldingnet,zhao20193d}, where the networks directly predict point locations in $\mathbb{R}^3$, our network $g_{w}(\cdot)$ models \textit{residuals} from the minimally-clothed body. 
To do so, we define a set of points $ \mathbf{t}_k\in \mathbb{R}^3$ on the posed body surface, and predict a local element (in the form of residuals) for each body point: $\mathbf{r}_{k,i}=g_{w}(\mathbf{u}_k, \mathbf{p}_i ; \bm{z}_k)$, where $\mathbf{p}_i$ denotes a sampled point from a local element.
In particular, an $\mathbf{r}_{k,i}$ is relative to a \textit{local coordinate system}\footnote{See the Appendix for more details on the definition of the local coordinates.} that is defined on $\mathbf{t}_k$.
To obtain a local element's position in the world coordinate $\mathbf{x}_{k,i}$, we apply articulations to $\mathbf{r}_{k,i}$ by the known transformation $\mathbf{T}_k$ associated with the local coordinate system, and add it to $\mathbf{t}_k$:
\begin{equation}
\label{eq:localcoord}
\mathbf{x}_{k,i} = \mathbf{T}_k \cdot \mathbf{r}_{k,i} + \mathbf{t}_k .
\end{equation}
Our network $g_{w}(\cdot)$ also predicts surface normals as an additional output for meshing and neural rendering, which are also transformed by $\mathbf{T}_k$. 
The residual formulation with explicit articulations is critical to clothed human modeling as the network $g_{w}(\cdot)$ can focus on learning local shape variations, which are roughly of the same scale. 
This leads to the successful recovery of fine-grained clothing deformations as shown in Sec.~\ref{sec:exp}. 
Next, we define local and global patch descriptors as well as the local shape feature \localfeatMath.

\paragraph{Local Descriptor.}
Each local element approximates a continuous small region on the target 2-manifold. Following \cite{yuan2018pcn}, we evenly sample $M$ points on a 2D grid and use them as a local patch descriptor: $\mathbf{p}=(p_i, q_i)\in \mathbb{R}^2$, with $~p_i, q_i\in [0,1],~i=1,2,\cdots,M$.
Within each surface element, all sampled points share the same global patch descriptor \descriptorMath and patch-wise feature \localfeatMath.

\paragraph{Global Descriptor.}
The global patch descriptor \descriptorMath in Eq.~\eqref{eq:scale} is the key to modeling different patches with a single network. While each global descriptor needs to be unique, it should also provide proximity information between surface elements to generate a globally coherent shape. Thus, we use 2D location on the UV positional map of the human body as a global patch descriptor: $\mathbf{u}_k = (u_k, v_k)$. While the 3D positions of a neutral human body can also be a global descriptor as in \cite{groueix20183d}, we did not observe any performance gain. Note that $\mathbf{T}_k$ and $\mathbf{t}_k$ in Eq.~\eqref{eq:scale} are assigned based on the corresponding 3D locations on the UV positional map.

\paragraph{Pose Embedding.}
To model realistic pose-dependent clothing deformations, we condition the proposed neural network with pose information from the underlying body as the local shape feature \localfeatMath.
While conditioning every surface element on global pose parameters $\bm{\theta}$ is possible, in the spirit of prior work~\cite{lahner2018deepwrinkles,CAPE:CVPR:20,patel20tailornet,yang2018analyzing}, we observe that such global pose conditioning does not generalize well to new poses and the network learns spurious correlations between body parts (a similar issue was observed and discussed in~\cite{STAR:ECCV:2020} for parametric human body modeling). Thus, we introduce a learning-based pose embedding using a 2D positional map, where each pixel consists of the 3D coordinates of a unique point on the underlying body mesh normalized by a transformation of the root joint. This 2D positional map is fed into a UNet~\cite{ronneberger2015u} to predict a 64-channel feature $\bm{z}_k$ for each pixel. The advantage of our learning-based pose embedding is two-fold: first, the influence of each body part is clothing-dependent and by training end-to-end, the learning-based embedding ensures that reconstruction fidelity is maximized adaptively for each outfit. Furthermore, 2D CNNs have an inductive bias to favor local information regardless of theoretical receptive fields~\cite{luo2016understanding}, effectively removing non-local spurious correlations. See Sec.~\ref{exp:clo_deform_local_global} for a comparison of our local pose embedding with its global counterparts.

\subsection{Training and Inference}\label{sec:losses}
For each input body, SCALE generates a point set $\mathbf{X}$ that consists of $K$ deformed surface elements, with $M$ points sampled from each element: $|\mathbf{X}|=KM$. From its corresponding clothed body surface, we sample a point set $\mathbf{Y}$ of size $N$ (i.e., $|\mathbf{Y}|=N$) as ground truth. 
The network is trained end-to-end with the following loss:
\begin{equation}\label{eq:total_loss}
\mathcal{L} = \lambda_\textrm{d}\mathcal{L}_\textrm{d}+ \lambda_\textrm{n}\mathcal{L}_\textrm{n}  + \lambda_\textrm{r}\mathcal{L}_\textrm{r} + \lambda_\textrm{c}\mathcal{L}_\textrm{c},
\end{equation}
where $\lambda_\textrm{d},\lambda_\textrm{n},\lambda_\textrm{r},\lambda_\textrm{c}$ are weights that balance the loss terms.

First, the Chamfer loss $\mathcal{L}_\textrm{d}$ penalizes bi-directional point-to-point $L2$ distances between the generated point set $\mathbf{X}$ and the ground truth point set $\mathbf{Y}$ as follows: 
\begin{equation}
\begin{aligned}\label{eq:chamfer}
\mathcal{L}_\textrm{d} = d(\mathbf{x},\mathbf{y}) =\frac{1}{KM} &\sum_{k=1}^K \sum_{i=1}^{M} \min _{j}\norm{\mathbf{x}_{k,i}-\mathbf{y}_j}_2^{2} \\
&+\frac{1}{N}\sum_{j=1}^N \min_{k,i}\norm{\mathbf{x}_{k,i}-\mathbf{y}_j}_2^{2}.
\end{aligned}
\end{equation}
For each predicted point $\mathbf{x}_{k,i}\in \mathbf{X}$, we penalize the $L1$ difference between its normal and that of its nearest neighbor from the ground truth point set: $\mathcal{L}_\textrm{n} = $
\begin{equation}\label{eq:normal_loss}
\frac{1}{KM}\sum_{k=1}^K \sum_{i=1}^{M}\bignorm{\bm{n}(\mathbf{x}_{k,i})
    - \bm{n}(\argmin_{\mathbf{y}_j\in\mathbf{Y}}d(\mathbf{x}_{k,i},\mathbf{y}_j))}_1,
\end{equation}
where $\bm{n}(\cdot)$ denotes the unit normal of the given point. We also add $L2$ regularization on the predicted residual vectors to prevent extreme deformations: 
\begin{equation}\label{eq:regularizer}
\mathcal{L}_\textrm{r} = \frac{1}{KM}\sum_{k=1}^K\sum_{i=1}^{M}\norm{\mathbf{r}_{k,i}}_2^2.
\end{equation}
When the ground-truth point clouds are textured, SCALE can also represent RGB color inference by predicting another $3$ channels, which can be trained with an $L1$ reconstruction loss:
\begin{equation}\label{eq:color_loss}
\mathcal{L}_\textrm{c} = \frac{1}{KM}\bignorm{\bm{c}(\mathbf{x}_{k,i})
    - \bm{c}(\argmin_{\mathbf{y}_j\in\mathbf{Y}}d(\mathbf{x}_{k,i},\mathbf{y}_j))}_1,
\end{equation}
where $\bm{c}(\cdot)$ represents the RGB values of the given point. 

\paragraph{Inference, Meshing, and Rendering.}
SCALE inherits the advantage of existing patch-based methods for fast inference. Within a surface element, we can sample arbitrarily dense points to obtain high-resolution point clouds. Based on the area of each patch, we adaptively sample points to keep the point density constant. Furthermore, since SCALE produces oriented point clouds with surface normals, we can apply off-the-shelf meshing methods such as Ball Pivoting~\cite{bernardini1999ball} and Poisson Surface Reconstruction (PSR)~\cite{kazhdan2006poisson,kazhdan2013screened}. As the aforementioned meshing methods are sensitive to hyperparameters, we present a method to directly render the SCALE outputs into high-resolution images by leveraging neural rendering based on point clouds~\cite{aliev2019neural,prokudin2020smplpix,Yifan:DSS:2019}. In Sec.~\ref{sec:exp}, we demonstrate that we can render the SCALE outputs using SMPLpix~\cite{prokudin2020smplpix}. See Appendix for more details on the adaptive point sampling and our neural rendering pipeline.

\section{Experiments}\label{sec:exp}
\subsection{Experimental Setup}
\paragraph{Baselines.} 
To evaluate the efficacy of SCALE's novel neural surface elements, we compare it to two state-of-the-art methods for clothed human modeling using meshes (CAPE~\cite{CAPE:CVPR:20}) and implicit surfaces (NASA~\cite{deng2019neural}). 
We also compare with prior work based on neural surface elements: AtlasNet~\cite{groueix2018atlasnet} and PCN~\cite{yuan2018pcn}. Note that we choose a minimally-clothed body with a neutral pose as a surface element for these approaches as in~\cite{groueix20183d} for fair comparison. To fully evaluate each technical contribution, we provide an ablation study that evaluates the use of explicit articulation, the global descriptor \descriptorMath, the learning-based pose embedding using UNet, and the joint-learning of surface normals.

\paragraph{Datasets.} We primarily use the CAPE dataset~\cite{CAPE:CVPR:20} for evaluation and comparison with the baseline methods. The dataset provides registered mesh pairs (clothed and minimally clothed body) of multiple humans in motion wearing common clothing (\eg T-shirts, trousers, and a blazer).
In the main paper we choose \textit{blazerlong} (blazer jacket, long trousers) and \textit{shortlong} (short T-shirt, long trousers) with subject \textit{03375} to illustrate the applicability of our approach to different clothing types. 
The numerical results on other CAPE subjects are provided in the Appendix. 
In addition, to evaluate the ability of SCALE to represent a topology that significantly deviates from the body mesh, we synthetically generate point clouds of a person wearing a skirt using physics-based simulation driven by the motion of the subject \textit{00134} in the CAPE dataset. 
The motion sequences are randomly split into training (70\%) and test (30\%) sets. 

\paragraph{Metrics.} We numerically evaluate the reconstruction quality of each method using Chamfer Distance (Eq.~\eqref{eq:chamfer}, in $m^2$) and the $L1$-norm of the unit normal discrepancy (Eq.~\eqref{eq:normal_loss}), evaluated over the $12,768$ points generated by our model.  For CAPE~\cite{CAPE:CVPR:20}, as the mesh resolution is relatively low, we uniformly sample the same number of points as our model on the surface using barycentric interpolation. As NASA~\cite{deng2019neural} infers an implicit surface, we extract an iso-surface using Marching Cubes~\cite{lorensen1987marching} with a sufficiently high resolution ($512^3$), and sample the surface.
We sample and compute the errors three times with different random seeds and report the average values.

\paragraph{Implementation details.} We use the SMPL~\cite{loper2015smpl} UV map with a resolution of $32\times32$ for our pose embedding, which yields $K=798$ body surface points (hence the number of surface elements). For each element, we sample $M=16$ square grid points, resulting in $12,768$ points in the final output. We uniformly sample $N=40,000$ points from each clothed body mesh as the target ground truth scan. More implementation details are provided in the Appendix.

\begin{table*}[tb]
    \centering
    \caption{\label{tab:exp_summary}\small Results of pose dependent clothing deformation prediction on unseen test sequences from the 3 prototypical garment types, of varying modeling difficulty.
    Best results are in \textbf{boldface}.}
    \small
    \begin{tabular}{p{5cm}p{1.15cm}<{\centering}p{1.05cm}<{\centering}p{1.05cm}<{\centering}p{1.05cm}<{\centering}p{1.05cm}<{\centering}p{1.05cm}<{\centering}p{2.35cm}<{\centering}}
    \toprule
      \multirow{2}{*}{Methods / Variants}  & \multicolumn{3}{c}{Chamfer-$L_2~(\times 10^{-4}m^2) \downarrow$} & \multicolumn{3}{c}{Normal diff. $(\times 10^{-1}) \downarrow$} &  \multirow{2}{*}{Inference time $(s) \downarrow$}\\
      \cmidrule{2-7}
      & \textit{blazerlong} & \textit{shortlong} & \textit{skirt}  & \textit{blazerlong} & \textit{shortlong} & \textit{skirt} &  \\
      \midrule
      \multicolumn{8}{c}{I. Comparison with SoTAs, Sec.~\ref{exp:clo_deform_nasa_cape}}\\
      \midrule
      CAPE~\cite{CAPE:CVPR:20} &  $1.96$& $1.37$ & -- & $1.28$& $1.15$  & -- & \textbf{0.013} \\
      NASA~\cite{deng2019neural} &  $1.37$ & $0.95$ & -- & $1.29$& $1.17$  & -- & 12.0\\
      Ours (SCALE, full model) &  \textbf{1.07} & \textbf{0.89} & \textbf{2.69} & \textbf{1.22} &  \textbf{1.12} & \textbf{0.94} & 0.009 (+ 1.1)\\
      \midrule
      \multicolumn{8}{c}{II. Global vs Local Elements, Sec.~\ref{exp:clo_deform_local_global}}\\
      \midrule
      a). global $\bm{z}$~\cite{groueix20183d} + AN~\cite{groueix2018atlasnet} & $6.20$ & $1.54$ & $4.59$ & $1.70$ & $1.86$ & $2.46$ & -- \\
      b). global $\bm{z}$~\cite{groueix20183d} + AN~\cite{groueix2018atlasnet} + Arti. & $1.32$ & $0.99$ & $2.95$ & $1.46$ & $1.23$ & $1.20$ & -- \\
      c). global $\bm{z}$~\cite{groueix20183d} + PCN~\cite{yuan2018pcn} + Arti. & $1.43$ & $1.09$ & $3.07$ &  $1.59$ & $1.40$ & $1.32$ & -- \\
      d). pose param $\bm{\theta}$ + PCN~\cite{yuan2018pcn} + Arti. & $1.46$ & $1.04$ & $2.90$ &  $1.61$ & $1.39$ & $1.33$  & -- \\
      \midrule
      \multicolumn{8}{c}{III. Ablation Study: Key Components, Sec.~\ref{exp:ablation}}\\
      \midrule
      e).~w/o Arti. $\mathbf{T}_k$ &  $1.92$ & $1.43$ & $2.71$ & $1.70$ & $1.53$ & $0.96$ & --\\
      f).~w/o \descriptorMath &  1.08 & \textbf{0.89} & 2.71 & \textbf{1.22} &  \textbf{1.12} & \textbf{0.94} & -- \\
      g).~UNet $\rightarrow$ PointNet, with \descriptorMath & $1.46$ & $1.03$ & $2.72$ & $1.34$ & $1.16$ & $0.95$ & -- \\
      h).~UNet $\rightarrow$ PointNet, w/o \descriptorMath & $1.92$& $1.42$ & $3.06$ & $1.69$ & $1.52$ & $1.39$ & -- \\
      i).~w/o Normal pred. & $1.08$ & $0.99$ & $2.72$ &  -- &  -- & -- & -- \\
     \bottomrule
    \end{tabular}
    \vspace{-1.35em}
\end{table*}

\begin{figure}[t]
    \centering
    \includegraphics[width=\linewidth]{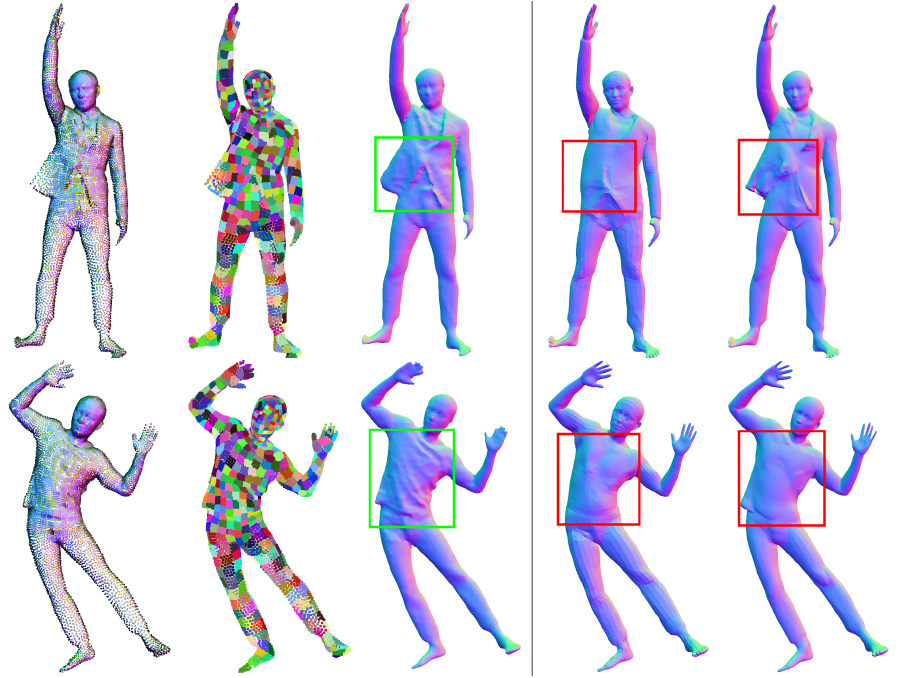}
    \put(-228,-10){\small{Ours\hspace{15pt}  Patch-colored \hspace{5pt} Meshed}}
    \put(-91,-10){\small{CAPE~\cite{CAPE:CVPR:20}}}
    \put(-43,-10){\small{NASA~\cite{deng2019neural}}}
    \caption{\small{\textbf{Qualitative comparison with mesh and implicit methods.} Our method produces coherent global shape, salient pose-dependent deformation, and sharp local geometry. The meshed results are acquired by applying PSR~\cite{kazhdan2013screened} to SCALE's point+normal prediction. The patch color visualization assigns a consistent set of colors to the patches, showing correspondence between the two bodies.}}
    \label{fig:pose_dpt_qualitative}
    \vspace{-10pt}
\end{figure}

\subsection{Comparison: Mesh and Implicit Surface}\label{exp:clo_deform_nasa_cape}
Block I of Tab.~\ref{tab:exp_summary} quantitatively compares the accuracy  and inference runtime of  SCALE, CAPE~\cite{CAPE:CVPR:20} and NASA~\cite{deng2019neural}. CAPE~\cite{CAPE:CVPR:20} learns the shape variation of articulated clothed humans as displacements from a minimally clothed body using MeshCNN~\cite{ranjan2018generatingcoma}. In contrast to ours, by construction of a mesh-based representation, CAPE requires registered templates to the scans for training. While NASA, on the other hand, learns the composition of articulated implicit functions without surface registration, it requires watertight meshes because the training requires ground-truth occupancy information. Note that these two approaches are unable to process the skirt sequences as the thin structure of the skirt is non-trivial to handle using the fixed topology of human bodies or implicit functions. 

For the other two clothing types, our approach not only achieves the best numerical result, but also qualitatively demonstrates globally coherent and highly detailed reconstruction results as shown in Fig.~\ref{fig:pose_dpt_qualitative}. 
On the contrary, the mesh-based approach~\cite{CAPE:CVPR:20} suffers from a lack of details and fidelity, especially in the presence of topological change. 
Despite its topological flexibility, the articulated implicit function~\cite{deng2019neural} is outperformed by our method by a large margin, especially on the more challenging \textit{blazerlong} data (22\% in Chamfer-$L_2$). 
This is mainly due to the artifacts caused by globally incoherent shape predictions for unseen poses, Fig.~\ref{fig:pose_dpt_qualitative}. 
We refer to the Appendix for extended qualitative comparison with the baselines. 

The run-time comparison illustrates the advantage of fast inference with the explicit shape representations. 
CAPE directly generates a mesh (with 7K vertices) in 13\textit{ms}. 
SCALE generates a set of 13K points within 9\textit{ms}; if a mesh output is desired, the PSR meshing takes 1.1\textit{s}. Note, however, that the SCALE outputs can directly be neural-rendered into images at interactive speed, see Sec.~\ref{exp:rendering}.
In contrast, NASA requires densely evaluating occupancy values over the space, taking 12\textit{s} to extract an explicit mesh.

\begin{figure*}[!tb]
    \centering
    \includegraphics[width=\linewidth]{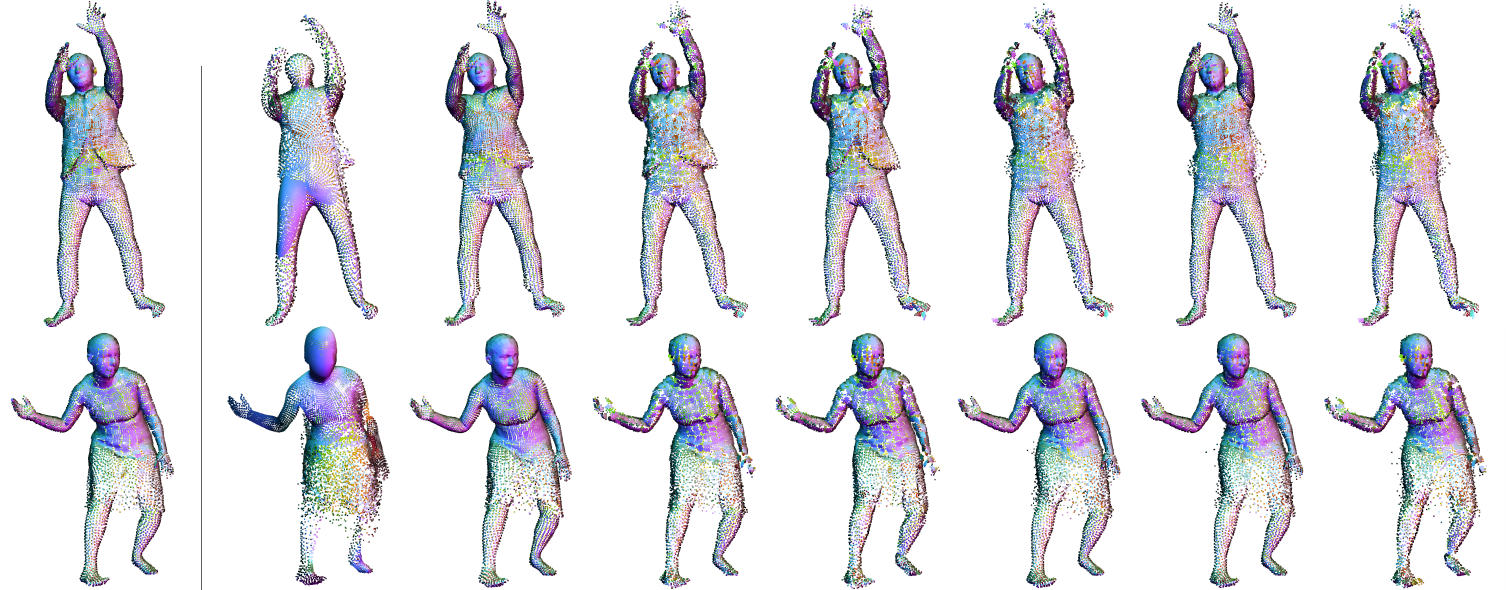}
    \put(-480, -12){\small{\hspace{4pt}Full model \hspace{32pt} $\bm{z}$ + AN \hspace{8pt} $\bm{z}$ + AN + Arti. \hspace{3pt} $\bm{z}$ + PCN + Arti.
    \hspace{1pt} $\bm{\theta}$ + PCN + Arti. \hspace{3pt}w/o Arti. $\mathbf{T}_k$}}
     \put(-100, -8){\small{PointNet \hspace{22pt} PointNet}}
     \put(-98, -16){\small{with \descriptorMath \hspace{28pt} w/o \descriptorMath }}
     \vspace{-2pt}
    \caption{\small{\textbf{Qualitative results of the ablation study.} Points are colored according to predicted normals. Our full model produces globally more coherent and locally more detailed results compared to the baselines. Note the difference at the bottom of the blazer (upper row) and the skirt (lower row).}}
    \label{fig:ablation_study}
    \vspace{-12pt}
\end{figure*}
\subsection{Global vs. Local Neural Surface Elements}\label{exp:clo_deform_local_global}
We compare existing neural surface representations~\cite{groueix20183d, groueix2018atlasnet, yuan2018pcn} in Fig.~\ref{fig:ablation_study} and block II of Tab.~\ref{tab:exp_summary}. Following the original implementation of AtlasNet~\cite{groueix2018atlasnet}, we use a global encoder that provides a global shape code $\bm{z} \in \mathbb{R}^{1024}$ based on PointNet~\cite{qi2017pointnet}. We also provide a variant of AtlasNet~\cite{groueix2018atlasnet} and PCN~\cite{yuan2018pcn}, where the networks predict residuals on top of the input body and then are articulated as in our approach. AtlasNet with the explicit articulation (b) significantly outperforms the original AtlasNet without articulation (a). This shows that our newly introduced articulated surface elements are highly effective for modeling articulated objects, regardless of neural surface formulations. As PCN also efficiently models a large number of local elements using a single network, (c) and (d) differ from our approach only in the use of a global shape code $\bm{z}$ instead of local shape codes. While (c) learns the global code in an end-to-end manner, (d) is given global pose parameters $\bm{\theta}$  {\em a priori}. Qualitatively, modeling local elements with a global shape code leads to noisier results. Numerically, our method outperforms both approaches, demonstrating the importance of modeling local shape codes. 
Notably, another advantage of modeling local shape codes is its parameter efficiency. The global approaches often require high dimensional latent codes (\eg 1024), leading to the high usage of network parameters (1.06M parameters for the networks above). In contrast, our local shape modeling allows us to efficiently model shape variations with significantly smaller latent codes (64 in SCALE) with nearly half the trainable parameters (0.57M) while achieving the state-of-the-art modeling accuracy. 

\subsection{Ablation Study}\label{exp:ablation}
We further evaluate our technical contributions via an ablation study. As demonstrated in Sec.~\ref{exp:clo_deform_local_global} and Tab.~\ref{tab:exp_summary}~(e), explicitly modeling articulation plays a critical role in the success of accurate clothed human modeling. We also observe a significant degradation by replacing our UNet-based pose embedding with PointNet, denoted as (g) and (h). This indicates that the learning-based pose embedding with a 2D CNN is more effective for local feature learning despite the conceptual similarity of these two architectures that incorporate spatial proximity information. Interestingly, the lack of a global descriptor derived from the UV map, denoted as (f), has little impact on numerical accuracy. As the similar ablation study between (g) and (h) shows significant improvement by adding $\mathbf{u}_k$ in the case of the PointNet architecture, this result implies that our UNet local encoder implicitly learns the global descriptor as part of the local codes \localfeatMath. As shown in Tab.~\ref{tab:exp_summary}~(i), another interesting observation is that the joint training of surface normals improves reconstruction accuracy, indicating that the multi-task learning of geometric features can be mutually beneficial. 

\begin{figure}[tb]
    \centering
    \includegraphics[width=0.98\linewidth]{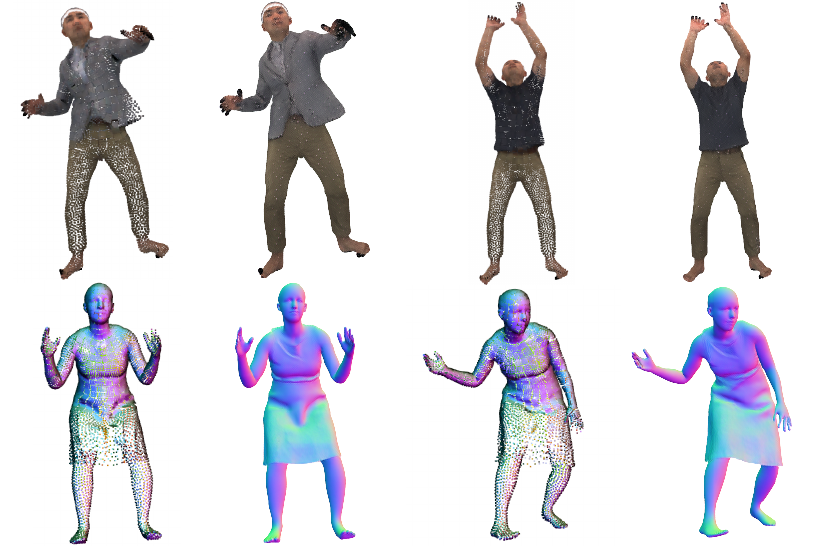}
    \caption{\small{\textbf{Neural Rendering of SCALE.} The dense point set of textured (upper row) or normal-colored (lower row) predictions from SCALE (the left image in each pair) can be rendered into realistic images with a state-of-the-art neural renderer~\cite{prokudin2020smplpix}.}}
    \vspace{-10pt}
    \label{fig:smplpix_rendering}
\end{figure}
\subsection{Neural Rendering of \app}\label{exp:rendering}
The meshing process is typically slow, prone to artifacts, and sensitive to the choice of hyperparameters.
To circumvent meshing while realistically completing missing regions, we show that generated point clouds can be directly rendered into high-resolution images with the help of the SMPLpix~\cite{prokudin2020smplpix} neural renderer, which can generate e.g.~a 512$\times$512 image in 42\textit{ms}. Fig.~ \ref{fig:smplpix_rendering} shows that the dense point clouds generated by SCALE are turned into complete images in which local details such as fingers and wrinkles are well preserved. Note that we show the normal color-coded renderings for the synthetic skirt examples, since they lack ground-truth texture information. 

\section{Conclusion}
We introduce SCALE, a highly flexible explicit 3D shape representation based on pose-aware local surface elements with articulation, which allows us to faithfully model a clothed human using point clouds without relying on a fixed-topology template, registered data, or watertight scans. 
The evaluation demonstrates that efficiently modeling a large number of local elements and incorporating explicit articulation are the key to unifying the learning of complex clothing deformations of various topologies. 

\paragraph{Limitations and future work.}
While the UV map builds a correspondence across all bodies, a certain patch produced by SCALE is not guaranteed to represent semantically the same region on the cloth in different poses. Jointly optimizing explicit correspondences~\cite{bhatnagar2020loopreg,PTF:CVPR:21} with explicit shape representations like ours remains challenging yet promising. 
Currently, SCALE models clothed humans in a subject-specific manner but our representation should support learning a unified model across multiple garment types.
While we show that it is possible to obviate the meshing step by using neural rendering, incorporating learnable triangulation \cite{liu2020ier, sharp2020ptn} would be useful for applications that need meshes.

\vspace{3pt}
\noindent
{\small 
{\bf   Acknowledgements}: We thank Sergey Prokudin for insightful discussions and the help with the SMPLpix rendering. 
Qianli Ma is partially funded by Deutsche Forschungsgemeinschaft (DFG, German Research Foundation) - 276693517 SFB 1233.

\noindent
{\bf   Disclosure:} MJB has received research gift funds from Adobe, Intel, Nvidia, Facebook, and Amazon. While MJB is a part-time employee of Amazon, his research was performed solely at, and funded solely by, Max Planck. MJB has financial interests in Amazon, Datagen Technologies, and Meshcapade GmbH.
}

\clearpage
\appendix
\balance
{\noindent\Large\textbf{Appendix}}
\setcounter{page}{1}
\counterwithin{figure}{section}
\counterwithin{table}{section}
\section{Implementation Details}
\subsection{Network Architectures}
\begin{figure}[!ht]
    \centering
    \includegraphics[width=\linewidth]{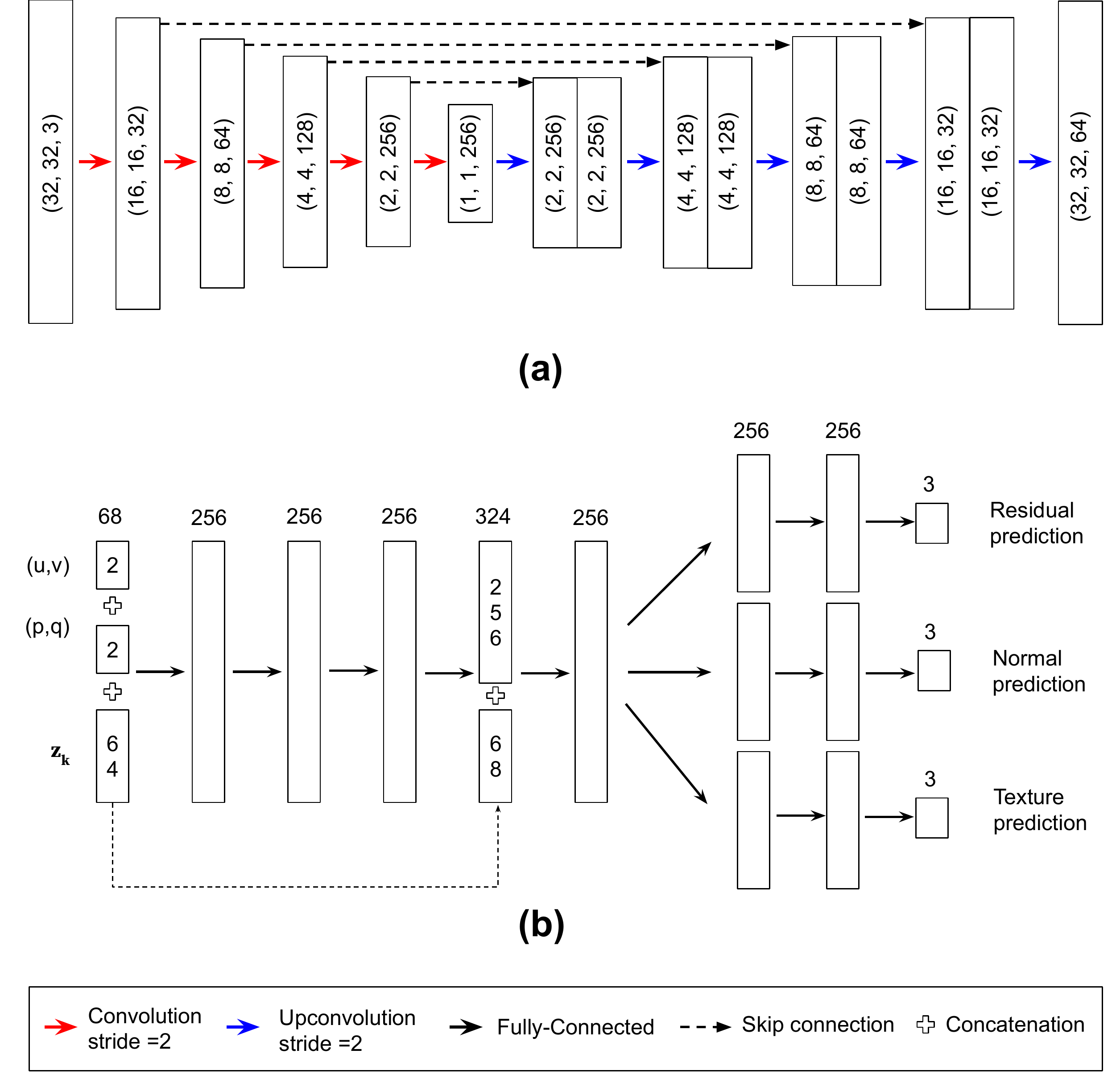}
    \caption{A visualization of our network architectures. \textbf{(a)} The UNet for our UV pose feature encoder. \textbf{(b)} The MLP for patch deformations. The numbers denote the dimensions of the network input or the layer outputs.}
    \label{fig:archi_detailed}
\end{figure}

To encode our UV positional map of resolution $32\times32$ into local features, we use a standard UNet~\cite{ronneberger2015u} as illustrated in Fig.~\ref{fig:archi_detailed}(a). It consists of five [Conv2d, BatchNorm, LeakyReLU(0.2)] blocks (red arrows), followed by five [ReLU, ConvTranspose2d, BatchNorm] blocks (blue arrows). The final layer does not apply BatchNorm.

To deform the local elements, we use an 8-layer MLP with a skip connection from the input to the 4th layer as in DeepSDF~\cite{park2019deepsdf}, see Fig.~\ref{fig:archi_detailed}(b). From the 6th layer, the network branches out three heads with the same architecture that predicts residuals from the basis point locations, normals and colors respectively.
Batch normalization and the SoftPlus nonlinearity with $\beta=1$ are applied for all but the last layer in the decoder.
The color prediction branch finishes with a Sigmoid activation to squeeze the predicted RGB values between 0 and 1. 
The predicted normals are normalized to unit length.

\subsection{Training and Inference}
We train SCALE with the Adam~\cite{kingma2014adam} optimizer with a learning rate of $3.0e-4$, a batch size of 16, for 800 epochs. As the early stage of the training does not reliably provide nearest neighbor points on the ground-truth, we add $\mathcal{L}_\textrm{n}$ and $\mathcal{L}_\textrm{c}$ when $\mathcal{L}_\textrm{d}$ roughly plateaus after 250 epochs. 

The residual, normal and color prediction modules are trained jointly. 
To balance the loss terms, the weights are set to $\lambda_\textrm{d}=2e4,\lambda_\textrm{r}=2e3,\lambda_\textrm{c}=\lambda_\textrm{n}=0$ at the beginning of the training, and $\lambda_\textrm{c}=\lambda_\textrm{n}=0.1$ from the 200th epoch when the point locations are roughly converged.

For the inference time comparison in the main paper Tab.~\ref{tab:exp_summary}, we report the wall-clock time using a desktop workstation with a Xeon CPU and Nvidia P5000 GPU. 

\subsection{Data Processing}
We normalize the bodies by removing the body translation and global orientation from the data. The motion sequences are randomly split into train (70\%) and test (30\%) sets. For the clothing types in the main paper, 
the number of train / test data samples is: \textit{blazerlong} 1334 / 563; \textit{shortlong} 3480 / 976; and \textit{skirt} 5113 / 2022.

\subsection{Definition of the Local Coordinates}
\begin{figure}[ht]
    \centering
    \includegraphics[width=0.9\linewidth]{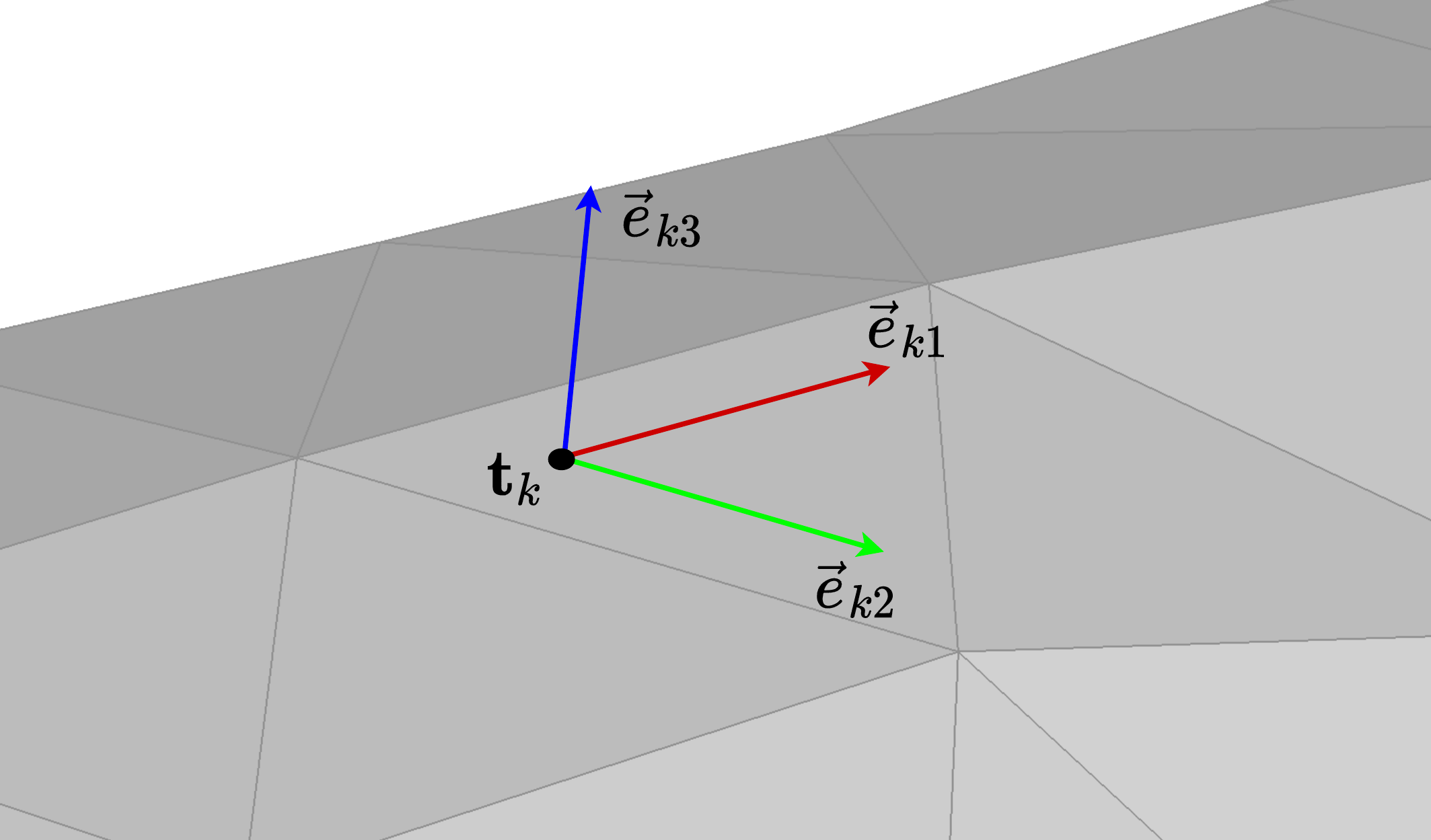}
    \caption{An illustration of the local coordinate system defined on a body point $\mathbf{t}_k$. We take the triangle where it locates on the SMPL body mesh (in grey), and build the local coordinate frame using the edges and surface normal of the triangle.}
    \label{fig:supp_localcoords}
\end{figure}

Here we elaborate on the local coordinate system used in the main paper Eq.~\eqref{eq:localcoord}. 
As illustrated in Fig.~\ref{fig:supp_localcoords}, for each body point $\mathbf{t}_k$, we find the triangle where $\mathbf{t}_k$ sits on the SMPL~\cite{loper2015smpl} body mesh. We take the first two edges $\vec{e}_{k1}, \vec{e}_{k2}$ of the triangle, as well as the normal vector of the triangle plane $\vec{e}_{k3}=\vec{e}_{k1}\times\vec{e}_{k2}$, as three axes of the local coordinate frame.
Note that $\vec{e}_{k1}, \vec{e}_{k2}, \vec{e}_{k3}$ are unit-length column vectors. 
The transformation associated with $\mathbf{t}_k$ is then defined as: $\mathbf{T}_k = [\vec{e}_{k1}, \vec{e}_{k2}, \vec{e}_{k3}]$.
The residual predictions $\mathbf{r}_k$ from the network are relative to the local coordinate system, and are transformed by $\mathbf{T}_k$ to the world coordinate according to the main paper Eq.~\eqref{eq:localcoord}.

\subsection{Adaptive Upsampling}
During inference, SCALE allows us to sample arbitrarily dense points to obtain high-resolution point sets. 
As the UV positional map provided by the SMPL model \cite{loper2015smpl} has a higher density around the head region and lower density around the legs, we mitigate the problem of unbalanced point density by resampling points proportional to the area of each local element. Note that we approximate the area of patches by summing the areas of triangulated local grid points. See Sec.~\ref{sec:adaptive_results} for qualitative results of the adaptive sampling.

\subsection{Neural Rendering}
Elaborating on the neural rendering of SCALE as shown in the main paper Sec.~\ref{exp:rendering}, we use the SMPLpix~\cite{prokudin2020smplpix} model for neural rendering. It takes as input an RGB-D projection of the colored point set generated by SCALE, and outputs a hole-filled, realistic, image of the predicted clothed human.

\paragraph{RGB-D projections.}
Given the colored point set, $ \mathbf{X}^+ = [\mathbf{X}, \mathbf{X}^c]\in\mathbb{R}^{KM\times6}$, where $\mathbf{X}^c\in\mathbb{R}^{KM\times3}$ are the RGB values of the points $\mathbf{X}$, we perform 2D projections using a pre-defined set of camera parameters $(\mathbf{K},\mathbf{R},\mathbf{t})$. 
The result is a set of RGB-D images, $I_\mathbf{x}\in\mathbb{R}^{W\times H\times 4}$. In the case where two points are projected to the same pixel, we take the value of the point that has smaller depth. These images are the inputs to the SMPLpix model.

\begin{figure}[t]
    \centering
    \includegraphics[width=0.97\linewidth]{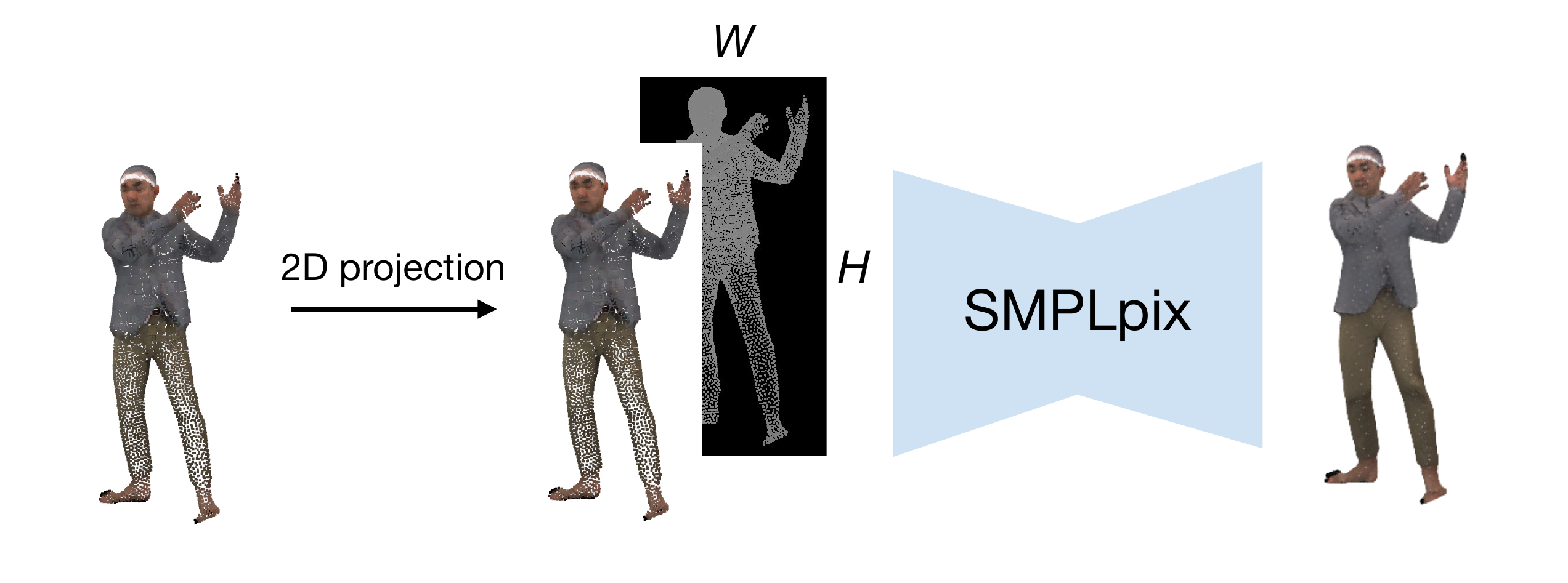}
    \put(-232,-6){\footnotesize{SCALE textured} \hspace{122pt} {Neural rendered}}
    \put(-150,-10){\footnotesize{RGB} \hspace{4pt} {Depth}}
    \put(-230,-16){\small{prediction(3D)} \hspace{140pt} {image}}
    \caption{Pipeline of neural rendering with SMPLpix~\cite{prokudin2020smplpix}.}
    \label{fig:smplpix_pipeline}
\end{figure}

\paragraph{Data and Training.} 
We train SMPLpix using the same data and train / test split as what we use to train SCALE.
Each (input, output) image pair for SMPLpix is acquired by performing the above-mentioned RGB-D projections to the SCALE predicted point set and the ground truth point set (of a higher density), respectively. 

Note that the distorted fingers or toes in some of our results stem from the artifacts present in the ground truth point clouds. Similarly, the holes in the ground truth scan data lead to occasional black color predictions on these regions. In addition, as the synthetic skirt data does not have ground truth texture, we use the point normals as the RGB values for the visualization and neural rendering.

The SMPLpix network is trained with the Adam optimizer~\cite{kingma2014adam} with a learning rate of $1e-4$, batch size 10, for 200 epochs, using the perceptual VGG-loss~\cite{johnson2016perceptual}.

\paragraph{Discussion.}
The neural point-based rendering circumvents the meshing step in traditional graphics pipelines.
Our SMPLpix implementation takes on average $42ms$ to generate a $512\times512$ image without any hardware-specific optimization. 
Recall that SCALE takes less than $9ms$ to generate a set of 13K points, our SCALE+SMPLpix pipeline remains highly efficient, and shows promise for future work on image-based clothed human synthesis with intuitive pose control.
Animations of the neural rendered SCALE results are provided in the supplemental video\footnote{Available at \url{https://qianlim.github.io/SCALE}.}.

\section{Additional Discussions}
\subsection{Tradeoff between Patches and Subsamples}
Experiments in the main paper use $K=798$ surface elements (which corresponds to a $32\times32$ UV positional map) and $M=16$ points per element.
In practice, these two numbers can be chosen per the specifications of the task. 

Here we discuss a degenerated case of our general formulation: $K=798\times16,~ M=1$. That is, we use a much higher number (12,768) of surface elements (corresponding to a $128\times128$ UV map), and sample only one example per element, whereby the number of output points remains the same.
Such a setting is equivalent to the traditional mesh vertex offset representation, where each body vertex corresponds to a point on the clothing surface.

We experiment with this setting (denoted as ``Vert-Offset'') and compare it to our method using surface elements in Tab.~\ref{tab:supp_tradeoff}. The number of the network parameters is kept the same for fair comparison.

The results reveal the advantage of our surface elements formulation: high fidelity and efficiency. 
Compared to the Vert-Offset representation, our method has $1/4$ FLOPS and lower GPU consumption in the UNet due to the $1/4$-sized UV map input. Nevertheless, it achieves overall comparable normal error and consistently lower Chamfer error.

\begin{table}[!h]
    \centering
    \caption{Comparison between our surface element formulation and vertex offset formulation.  Chamfer-$L_2\times10^{-4}m^2$, normal diff $\times10^{-1}$.}
    \begin{tabular}{llp{1.6cm}<{\centering}p{1.6cm}<{\centering}}
    \toprule
    & & Vert-Offset & Ours \\
    \midrule
    \multirow{3}{*}{Chamfer-$L_2$}  &\textit{blazerlong} & 1.13 &  \textbf{1.07}\\
    &\textit{shortlong}& 0.91 & \textbf{0.89}\\
    &\textit{skirt}& 2.78 & \textbf{2.69}\\
    \midrule
    \multirow{3}{*}{Normal diff} &\textit{blazerlong}& \textbf{1.20} & 1.22 \\
    &\textit{shortlong}& \textbf{1.09} & 1.12 \\
    &\textit{skirt}& 0.97 & \textbf{0.94}\\
    \bottomrule
    \end{tabular}
    \label{tab:supp_tradeoff}
\end{table}

\subsection{Effect of Adaptive Upsampling}\label{sec:adaptive_results}
\begin{figure}[tb]
    \centering
    \includegraphics[width=\linewidth]{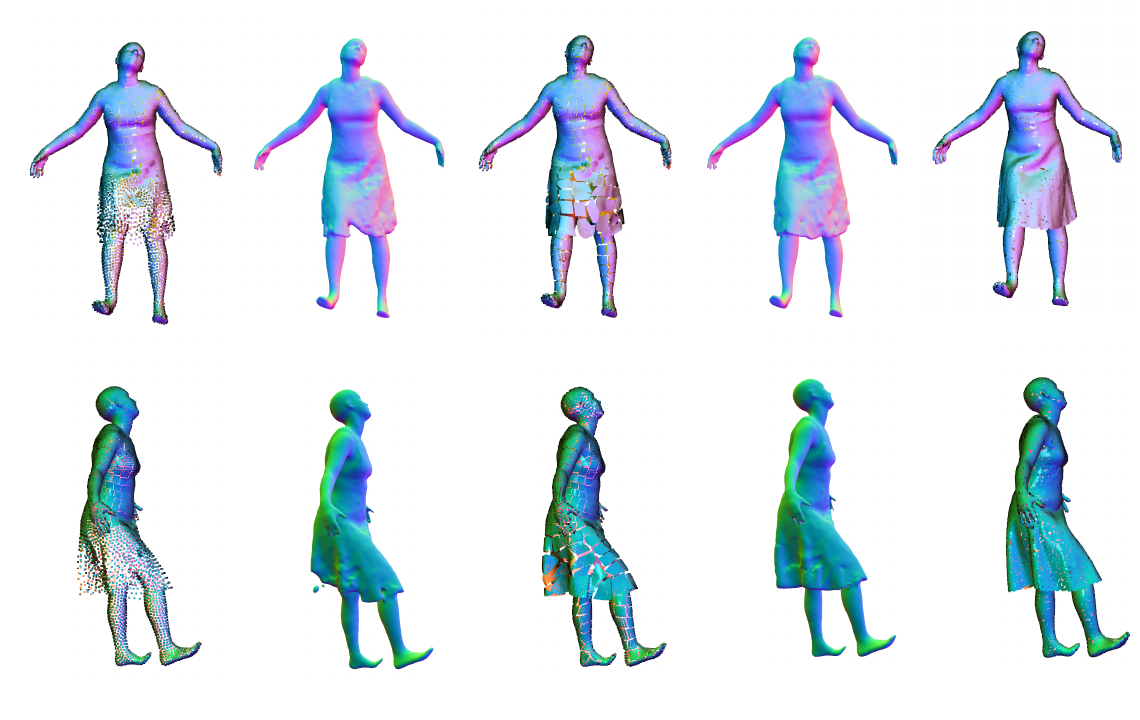}
    \put(-235,-10){\small{w/o adaptive} \hspace{2pt} {w/o adaptive} \hspace{2pt} {w/  adaptive}\hspace{2pt} {w/  adaptive}}
    \put(-20,-16){\small{GT}}
    \put(-224,-21){\small{points} \hspace{22pt} {meshed} \hspace{22pt} {points} \hspace{18pt} {meshed}}

    \caption{Qualitative effects of the adaptive patch subsampling.}
    \label{fig:adaptive_qualitative}
\end{figure}

Fig.~\ref{fig:adaptive_qualitative} shows the effect of adaptive patch sampling at test time. Due to the unbalanced point density in the SMPL UV map, the SCALE output will have sparser points on the leg region if the same number of points are sampled for every surface element. 
Such sparse points can be insufficient to represent the complex garment geometry in these regions, \eg in the case of skirts. 
Consequently, when applying Poisson Surface Reconstruction~\cite{kazhdan2013screened} to them, the reconstructed mesh will have missing geometry and ghosting artifacts, as demonstrated in the second column of Fig.~\ref{fig:adaptive_qualitative}. 
Adaptive patch upsampling adds more points to the bigger patches. With more points sampled on the skirt surface, the mesh reconstruction quality is improved.

The figure also shows a limitation of our model: once the model is trained, the test-time adaptive upsampling can only increase the point density \textit{within} each patch, while the gap between patches cannot be shrunk. As discussed in the main paper, a potential solution is to more explicitly model the connectivity between the patches by incorporating a learnable triangulation. We leave this as future work.

\subsection{Extended Result Analysis}
\paragraph{Error analysis on the patch periphery.} 
From our qualitative results (main paper Figs.~\ref{fig:pose_dpt_qualitative},\ref{fig:ablation_study},\ref{fig:smplpix_rendering}), the patches can sufficiently deform to represent fine structures such as fingers. Here, we perform additional numerical analysis by calculating the mean single-directional Chamfer error (from the predicted points to the ground truth points) of the patches' peripheral points and inner points respectively. 
We observe a slightly larger Chamfer error (4\%) from the peripheral points than the patch center points. The low relative difference between the two is in line with the qualitative results, yet shows space for improvement in future work.

\begin{figure}[!h]
    \centering
    \includegraphics[width=0.75\linewidth]{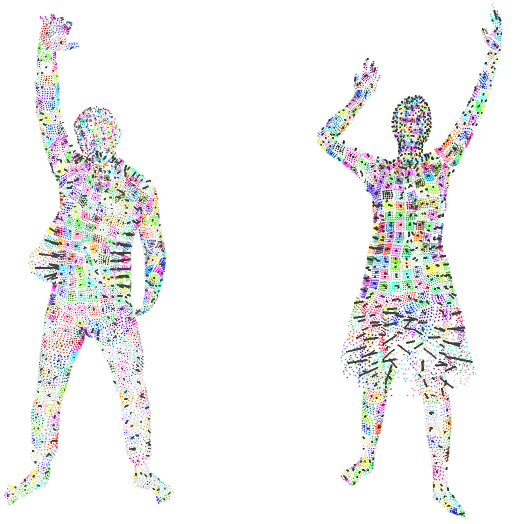}
    \caption{An illustration of the correspondence between the body points and the patches. Each black line connects a body basis point and the center of its corresponding patch.}
    \label{fig:correspondence}
\end{figure}

\paragraph{Clothing-body correspondence.} Fig.~\ref{fig:correspondence} illustrates the correspondence between the patches on the clothed body surface and the basis points from the underlying body.

\subsection{Additional Evaluations}
\paragraph{CAPE data.}
In the main paper, we highlight the characteristics of our model on the two prototypical outfit types (\textit{blazerlong} and \textit{shortlong}) from the CAPE dataset. Here we show the results on the rest of the CAPE dataset, which comprises mostly tight-fitting clothes such as short / long T-shirts, dress shirts, and short / long trousers. For each outfit, $30\%$ of the sequences are selected for testing and the rest are for training. 

\begin{table}[h]
    \centering
    \caption{Reconstruction error on the entire CAPE dataset.  Chamfer-$L_2\times10^{-4}m^2$, normal diff $\times10^{-1}$.}
    \label{tab:quant_full_cape}
    \begin{tabular}{lccc}
    \toprule
     & CAPE~\cite{CAPE:CVPR:20} & NASA~\cite{deng2019neural} &  Ours \\
    \midrule
    Chamfer-$L_2$ & 1.28 & 4.08 & \textbf{0.93} \\
    Normal diff & \textbf{1.16} & 1.24 & 1.18\\
    \bottomrule
    \end{tabular}
\end{table}

\begin{figure}[h]
    \centering
    \includegraphics[width=\linewidth]{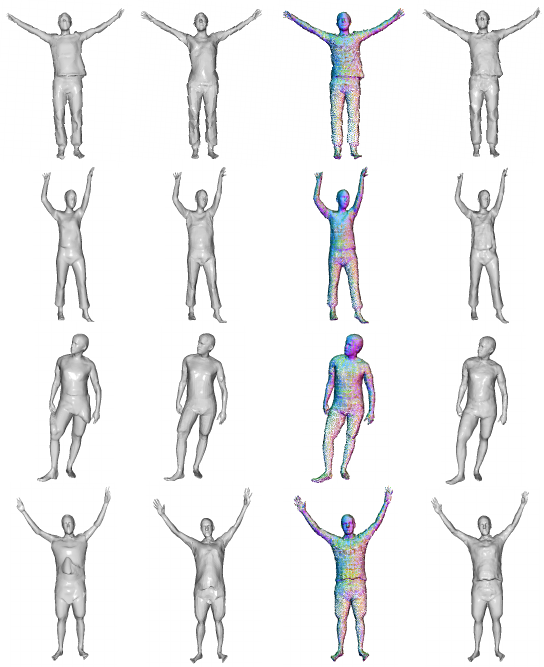}
    \put(-224,305){NASA~\cite{deng2019neural}~~ CAPE~\cite{CAPE:CVPR:20}}
    \put(-106,310){SCALE\qquad~~~ SCALE}
    \put(-118,300){Points+Normals\quad Meshed}
    \caption{Extended qualitative results on the CAPE dataset.}
    \label{fig:more_qualitative}
\end{figure}

As shown in Tab.~\ref{tab:quant_full_cape}, our model again outperforms both baselines in the Chamfer error by a large margin, and is comparable with the CAPE model in terms of normal accuracy. 
Note that on several sequences NASA predicts bodies with missing limbs, hence the high Chamfer error.
Qualitative results in Fig.~\ref{fig:more_qualitative} are also in accordance with the main paper experiments. Results from the CAPE model~\cite{CAPE:CVPR:20} in general lack realistic pose-dependent clothing deformation.
NASA~\cite{deng2019neural} can predict detailed clothing structure with notable influences of the body pose, but often suffer from discontinuities between different body parts. 
SCALE produces clothing shapes that naturally move with varied poses, showing a coherent global shape and detailed local structures such as wrinkles and edges.

\begin{table*}
    \centering
    \caption{Comparison between our method with methods that use a global feature code on the long dress data. ``+Arti.'' denotes applying articulation. Chamfer-$L_2\times10^{-4}m^2$, normal diff $\times10^{-1}$.}
    \label{tab:quant_long_dress}
    \begin{tabular}{lccccc}
    \toprule
     & Global $\bm{z}$+& Global $\bm{z}$+& Global $\bm{z}$+& Pose params+& SCALE \\
     & AtlasNet~\cite{groueix2018atlasnet} & AtlasNet~\cite{groueix2018atlasnet}+Arti. & PCN~\cite{yuan2018pcn}+Arti. & PCN~\cite{yuan2018pcn}+Arti. & (Ours) \\
    \midrule
    Chamfer-$L_2$ & 16.03 & 10.47 & 8.88 & 8.69 & \textbf{8.41} \\
    Normal diff & 3.00 & 1.62 & 1.70 & 1.70 & \textbf{1.32} \\
    \bottomrule
    \end{tabular}
\end{table*}

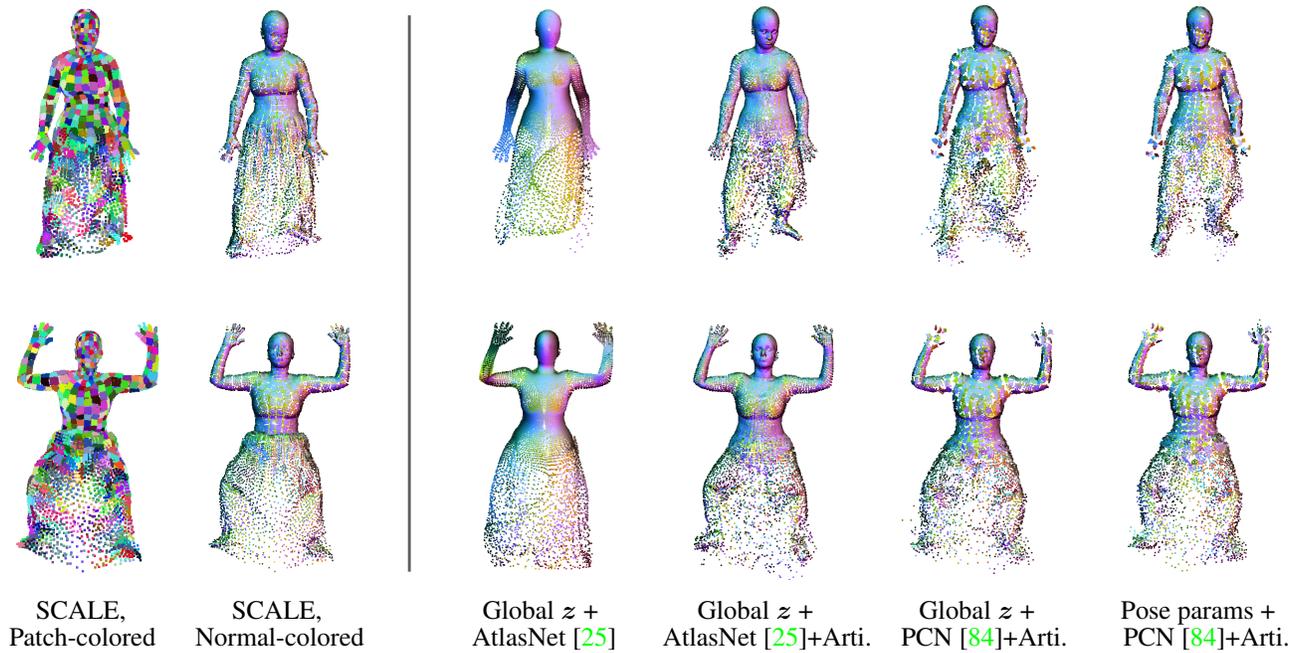
\begin{figure*}
    \centering
    \includegraphics[width=\linewidth]{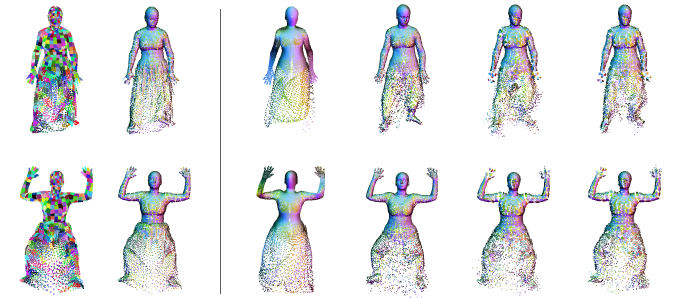}
    \put(-475,-15){SCALE,\qquad\qquad SCALE,}
    \put(-485,-25){Patch-colored~~\quad Normal-colored}
    \put(-306,-15){Global $\bm{z}$ + \qquad\quad~ Global $\bm{z}$ + \quad\qquad~~ Global $\bm{z}$ + \qquad\quad Pose params +}
    \put(-310,-25){AtlasNet~\cite{groueix2018atlasnet} \quad~ AtlasNet~\cite{groueix2018atlasnet}+Arti. ~~ PCN~\cite{yuan2018pcn}+Arti. \quad~~ PCN~\cite{yuan2018pcn}+Arti.}
    \vspace{6pt}
    \caption{Qualitative results on the long dress data. ``+Arti.'' denotes applying articulation.}
    \label{fig:long_dress}
\end{figure*}

\paragraph{Long dress.} In addition to the mid-length skirt in the main paper, we also experiment with a more challenging long dress. Similar to the skirt, the long dress data are created with physics-based simulation. Since the dress deviates from the body topology and contains thin cloth structures, both baselines, CAPE and NASA, are unable to process it. Here we compare with methods that use a global feature code with the same setting as in the main paper Sec.~\ref{exp:clo_deform_local_global}.

As shown in Tab.~\ref{tab:quant_long_dress} and Fig.~\ref{fig:long_dress}, for either using a large global surface element (as in AtlasNet~\cite{groueix2018atlasnet} and 3D-CODED~\cite{groueix20183d}, the first two columns in Tab.~\ref{tab:quant_long_dress}) or numerous local surface elements (as in PCN~\cite{yuan2018pcn}, the last two columns in Tab.~\ref{tab:quant_long_dress}), decoding the shape from a global shape code in general fails to reconstruct the clothing geometry faithfully, resulting in high numerical errors. In contrast, SCALE is able to represent the wrinkles and folds on the dress while producing a smooth shape for the upper body, validating our key design choices, i.e.~local feature codes and explicit articulation modeling.

\subsection{Animated Results}
Please refer to the supplemental video at {\small\url{https://qianlim.github.io/SCALE}} for more qualitative comparisons against CAPE and NASA, as well as animated results produced by SCALE.

\clearpage

{\small
\bibliographystyle{ieee_fullname}
\bibliography{references}
}

\end{document}